\documentclass[letterpaper]{article} 
\usepackage{aaai25}  
\usepackage{times}  
\usepackage{helvet}  
\usepackage{courier}  
\usepackage[hyphens]{url}  
\usepackage{graphicx} 
\usepackage{natbib}  
\usepackage{caption} 
\usepackage{algorithm}
\usepackage{algorithmic}

\usepackage{newfloat}
\usepackage{listings}
\DeclareCaptionStyle{ruled}{labelfont=normalfont,labelsep=colon,strut=off} 
\floatstyle{ruled}
\newfloat{listing}{tb}{lst}{}
\floatname{listing}{Listing}
\usepackage{subcaption}
\usepackage{booktabs}
\usepackage{amssymb}
\usepackage{amsmath}
\usepackage{multirow}
\usepackage{tabularx}
\usepackage{multirow}
\usepackage{makecell}
\usepackage{subcaption}
\usepackage[labelfont=bf]{caption}
\usepackage{dblfloatfix}
\usepackage{tabularray}
\usepackage{rotating}
\usepackage{bm}

\title{WPMixer: Efficient Multi-Resolution Mixing for Long-Term Time Series Forecasting}

\author {
    Md Mahmuddun Nabi Murad\textsuperscript{\rm 1},
    Mehmet Aktukmak\textsuperscript{\rm 2},
    Yasin Yilmaz\textsuperscript{\rm 1}
}
\affiliations {
    \textsuperscript{\rm 1}University of South Florida\\
    \textsuperscript{\rm 2}Intel Corporation\\
    mmurad@usf.edu, mehmet.aktukmak@intel.com, yasiny@usf.edu
}

\begin{document}
\setlength{\tabcolsep}{1mm} 
\maketitle

\begin{abstract}
Time series forecasting is crucial for various applications, such as weather forecasting, power load forecasting, and financial analysis. In recent studies, MLP-mixer models for time series forecasting have been shown as a promising alternative to transformer-based models. However, the performance of these models is still yet to reach its potential. In this paper, we propose Wavelet Patch Mixer (WPMixer), a novel MLP-based model, for long-term time series forecasting, which leverages the benefits of patching, multi-resolution wavelet decomposition, and mixing. Our model is based on three key components: (i) multi-resolution wavelet decomposition, (ii) patching and embedding, and (iii) MLP mixing. Multi-resolution wavelet decomposition efficiently extracts information in both the frequency and time domains. Patching allows the model to capture an extended history with a look-back window and enhances capturing local information while MLP mixing incorporates global information. Our model significantly outperforms state-of-the-art MLP-based and transformer-based models for long-term time series forecasting in a computationally efficient way, demonstrating its efficacy and potential for practical applications.
\end{abstract}

\section{Introduction}
Typically, time series data volume accumulates to vast amounts in various applications due to recording observations and events over long time horizons. The study of predicting time series data has been essential because of its extensive use in various domains such as finance, weather forecasting, and energy consumption prediction.

While research in time-series forecasting, for a long time, relied on traditional statistical methods such as ARIMA \cite{c:arima}, HMM \cite{c:hmm}, and SSM \cite{c:ssm}, with the increasing availability of large datasets and high computational power, deep learning methods gained prevalence due to their superior performance in complex tasks. Specifically, RNN and CNN-based models like DeepAR \cite{c:deepar}, and SCINet \cite{c:scinet}, as well as transformer-based time series forecasting models, have become popular over time. 

Transformer models for time series forecasting, such as Informer \cite{c:informer}, Autoformer \cite{c:autoformer}, Fedformer \cite{c:fedformer}, and Crossformer \cite{c:crossformer} have become popular thanks to their improved capability of learning long-term dependencies. However, recently, questions have arisen about the performance of the transformer variants in time series forecasting. The study \cite{c:dlinear} demonstrated that a simple linear model can outperform or perform similarly with the state-of-the-art transformers on the popular benchmark datasets for time series forecasting. 

Recently, MLP-based models have outperformed transformer variants in this domain. TimeMixer \cite{c:timemixer} and TSMixer \cite{c:tsmixer} showed excellent prospects in multivariate time series forecasting. TSMixer, an MLP-mixer-based variant, mixes data in the time and channel domain but is computationally expensive for long-term forecasting due to a longer look-back window. TimeMixer, which achieves the state-of-the-art results on most benchmark datasets, decomposes a multi-scaled time series into seasonal and trend series using the moving average method and then employs the mixing among the mult-scaled data. However, due to complex seasonality patterns, decomposing a signal into seasonal and trend data is inadequate, and mixing among the multi-scaled data can cause information loss \cite{c:hierarchicaltimeseries}. Additionally, real-world time series data can have abrupt spikes and dips, which is difficult to explain using multi-scaled moving average-based decomposition techniques. Furthermore, capturing the information only in the time domain is not sufficient due to the complex nature of the time series data. SWformer, a variant of Sepformer \cite{c:sepformer}, extracts information in the time and frequency domain utilizing wavelet transform-based decomposition. However, a multi-level wavelet transform is required to achieve its full potential.

To address these challenges, we propose a novel MLP-mixer-based model, called Wavelet Patch Mixer (WPMixer). What sets our model apart is its ability to capture intricate information in both the time and frequency domains, achieved through the use of multi-level wavelet decomposition. WPMixer decomposes the time series into multiple approximation and detail coefficient series using the multi-level wavelet transform. Distinct resolution branches handle each coefficient series, preventing information loss from mixing among multiple coefficient series. We utilize patching to capture local information and reduce the computational cost. We also employ patch mixer followed by embedding mixer to capture global information. Our contributions can be summarized as follows:
\begin{itemize}
\item We propose a novel model consisting of three core parts. Multi-level wavelet decomposition enables utilizing time and frequency domain properties due to spikes and dips, which cannot be captured by moving average-based decomposition methods in the time domain. Patching and mixing, on the other hand, capture local and global information, respectively. 
\item We analyze each decomposed series using a distinct resolution branch. This approach ensures that information from each resolution is maintained separately, thereby minimizing potential information loss.
\item We enhance the performance of the patch mixer by applying an embedding mixer after each patch mixer. 
\item Our model, WPMixer, efficiently achieves state-of-the-art performance in long-term forecasting on several benchmark datasets. 
\end{itemize}


\section{Related Works}
\label{Relatedworks}
Time series forecasting refers to predicting a sequence of values in a time series based on a past sequence. Research on time series forecasting considers both long-term and short-term forecasting tasks.

Transformer-based models have recently shown remarkable performance in long-term forecasting. Informer \cite{c:informer} applies prob-sparse attention with distill operation. Autoformer \cite{c:autoformer} improves Informer by applying decomposition in the transformer architecture. They decompose time series into seasonal and trend patterns with auto-correlation mechanisms based on time series periodicity. Sepformer \cite{c:sepformer} and FEDformer \cite{c:fedformer} are other transformer models which use decomposition techniques for long-term time series forecasting. Sepformer uses a single-level wavelet decomposition, in which wavelet coefficients are processed by a transformer. FEDformer enhances the time domain features using Fourier and wavelet transforms. In addition to the enhancement method, they also utilize separate attention mechanisms for Fourier and wavelet decomposed data. The Crossformer \cite{c:crossformer} model employes a dual-stage attention mechanism to capture dependencies across time and variables. In \cite{c:stationary}, a non-stationary transformer is proposed with de-stationary attention to address the over-stationarization problem. In the framework of PatchTST \cite{c:patchtst}, a conventional transformer augmented with patching is introduced to address the challenge of minimizing computational complexity while effectively capturing local semantic information. iTransformer \cite{c:itransformer}, an exclusively encoder-based transformer architecture, adopts a strategy of tokenizing each variate series individually rather than processing multivariate data at a single time step. This approach facilitates the computation of mutual attention across the multivariate series.

FiLM \cite{c:film} modifies the time series by transforming it into a Legendre polynomial space, thereby preserving the memory of long-term historical data. This method employs a frequency-enhanced operation akin to that used by FEDFormer \cite{c:fedformer} to accomplish the enhancement of the time series data. MICN \cite{c:micn} employs multiscale hybrid decomposition to analyze seasonal and trend components. Forecasting seasonal series is conducted using a convolutional neural network (CNN) model, which implements a convolutional kernel in the time domain. Trend prediction is achieved through a regression-based approach. TimesNet \cite{c:timesnet} utilizes the Fast Fourier transform to derive multiple periods for transforming time series data, thereby elucidating inter-period and intra-period variations within the series. In \cite{c:dlinear}, authors presents a group of linear models to demonstrate the effectiveness of simple linear models against the transformer-based model.

Recently, MLP-Mixer models have also been shown to provide effective solutions for time series forecasting despite being initially proposed for vision-based tasks \cite{c:mlpmixervision}. This potential is further demonstrated in TSMixer \cite{c:tsmixer} and TimeMixer \cite{c:timemixer}, where the mixer model is shown to outperform the transformer-based methods on the popular benchmark datasets. TSMixer has the same architecture as the original MLP-Mixer \cite{c:mlpmixervision}, but instead of mixing in the patch and channel domain, it mixes data in the time and channel domain directly. TimeMixer obtains a multi-scaled time series by applying down-sampling, then decomposes the multi-scaled time series into seasonal and trend series and mixes the data. 

In our proposed WPMixer model, we improve the performance of the MLP-mixer-based models by employing multi-level wavelet decomposition with patching and mixing. 

\section{Proposed Method}
\label{Proposedmethod}
Given a multivariate time series $\bm{X}_L = \{\bm{x}_{t-L+1},\ldots, \bm{x}_{t-1}, \bm{x}_t\}$, with a look-back window $L$, at time step $t$, we aim to forecast the subsequent $T$ data points $\bm{X}_T = \{\bm{x}_{t+1}, \bm{x}_{t+2},\ldots, \bm{x}_{t+T}\}$, where $\bm{x}_t\in\mathbb{R}^{1\times C}$ denotes a multivariate data point at time $t$, $C$ is the number of the variates, and $T$ is the prediction length.

\begin{figure*}[!htb]
    \centering
    \includegraphics[width=\textwidth, height=0.6\textheight, keepaspectratio]{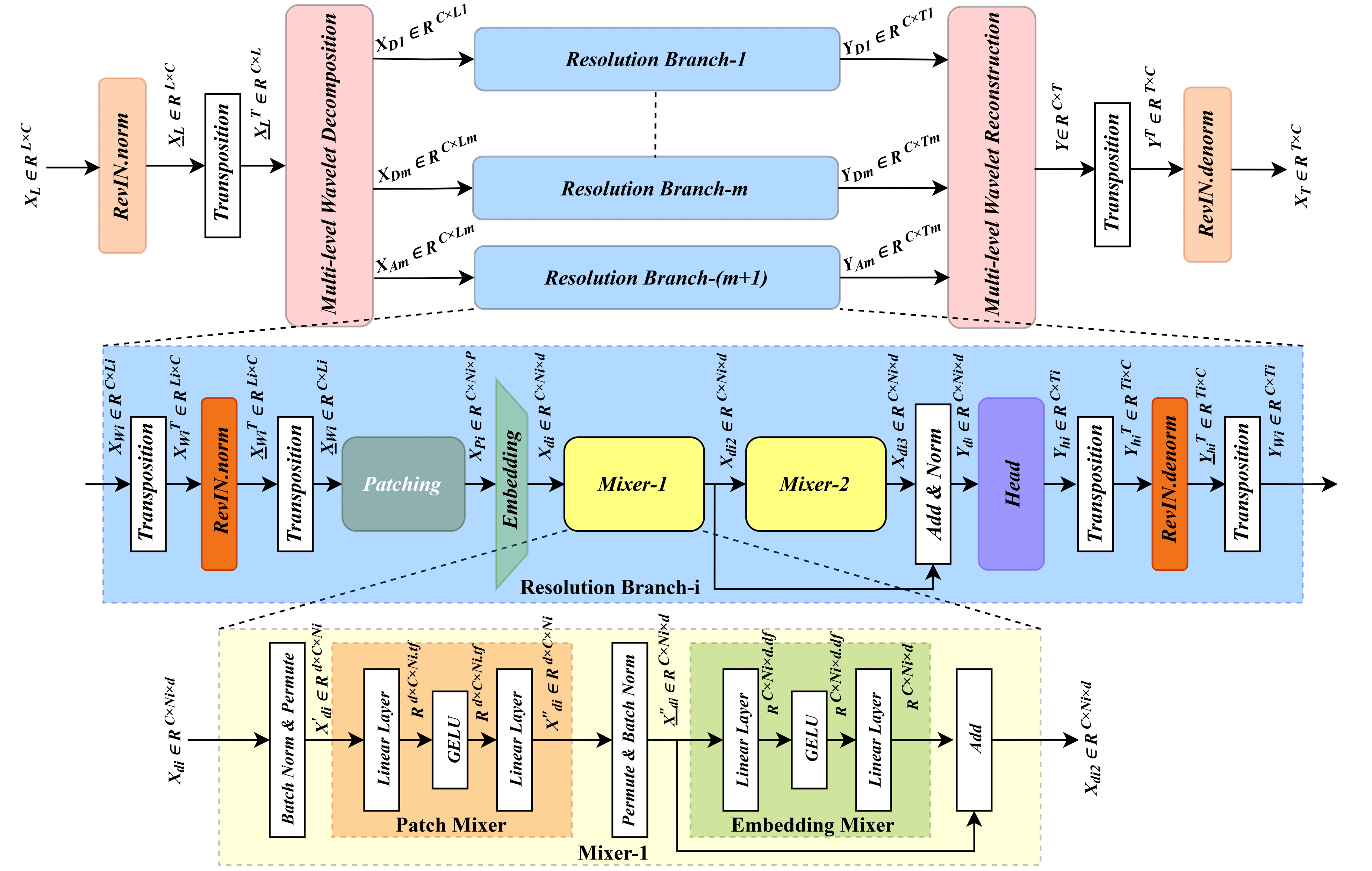}
    \caption{WPMixer with $m$ levels of wavelet decomposition. $\bm{X}_{A_i}$ and $\bm{X}_{D_i}$ 
    are the approximation and detail coefficient series corresponding to the input time series $\bm{X}_{L}$. $\bm{Y}_{A_i}$ and $\bm{Y}_{D_i}$ are the predicted approximation and detail coefficient series corresponding to the predicted time series $\bm{X}_{T}$. To simplify notation, $\bm{X}_{W_i}$ denotes either $\bm{X}_{A_i}$ or $\bm{X}_{D_i}$. Code is available at \url{https://github.com/Secure-and-Intelligent-Systems-Lab}}
    \label{fig:overall_model_architecture}
\end{figure*}

\subsection{Model Architecture}
The architecture of the proposed model is illustrated in Figure \ref{fig:overall_model_architecture}. Our approach begins with decomposing the normalized time series data into approximation and detail coefficient series through multi-level wavelet decomposition. This multi-level decomposition facilitates feature extraction from the time series data at various resolutions, where each resolution represents a distinct frequency level. As we progress to higher decomposition levels, the frequency range of the approximation coefficients becomes narrower. At the same time, we get multiple detail coefficient series that represent detailed information at various frequency levels. However, higher-level coefficient series may not always yield relevant information for forecasting tasks. Additionally, different wavelets offer varying trade-offs between time and frequency localization, making the selection of an optimal decomposition level and wavelet type a crucial aspect of the optimization process.

Our model processes each wavelet coefficient series through a distinct resolution branch, which prevents the intermixing of information across different frequency scales. Each resolution branch comprises an instance normalization module, a patch and embedding module, several mixer modules, a head module, and an instance denormalization module. The patch and embedding module transforms the normalized wavelet coefficient series into a series of patches. The patch mixer modules then aggregate the local information contained within these patches into a global information context. In the mixer module, which is a fusion of a patch mixer and an embedding mixer, the embedding mixer captures the global information in a higher dimensional space. The head module subsequently forecasts the wavelet coefficient series, providing the information needed for predicting the time series. A denormalization layer is employed to reintegrate the stationary information into the predicted wavelet coefficient series. Finally, the multi-level wavelet reconstruction module reconstructs the predicted time series by utilizing the predicted approximation and detail wavelet coefficient series. In the following subsections, we describe the key modules of our model.

\subsubsection{Instance Normalization:}
One of the main challenges for time series forecasting is to deal with the time-varying mean and variation. To overcome this challenge, Reversible Instance Normalization (RevIN) with learnable affine transform has been proposed in \cite{c:revin}. 
We initially employ RevIN normalization and denormalization directly in the time series data before decomposition and after reconstruction, respectively. We also employ RevIN normalization and denormalization in the wavelet coefficient series. The positions of the RevIN normalization and denormalization layers are shown in Fig \ref{fig:overall_model_architecture}. 

\subsubsection{Decomposition:}
We utilize the multi-level discrete wavelet transform to decompose the time series data. This transformation involves an iterative decomposition process utilizing high-pass and low-pass filters to extract wavelet coefficients at multiple levels \cite{c:mallat}. The coefficients of the filters depend on the type of wavelet. The output of the high-pass filter refers to detailed information, called detail coefficients, whereas the output of the low-pass filter refers to low-frequency information, called approximation coefficients. At each level, the approximation coefficients from the preceding level is split into new approximation and detail coefficients, allowing for a deeper data analysis. We adapt the implementation of the multi-level discrete wavelet transform from \cite{c:19} to work with PyTorch mixed precision analysis. \\ 

The decomposition module disintegrates the normalized time series $\underline{\bm{X}}_{L}^{T} \in \mathbb{R}^{C \times L}$ into approximation and detail coefficient series: 
\begin{equation}
[\bm{X}_{A_m}, \bm{X}_{D_m}, \bm{X}_{D_{m-1}}, \ldots, \bm{X}_{D_1}] = Decomp(\underline{\bm{X}}_{L}^{T} , \psi, m).
\label{eq:pmd1}
\end{equation}
In this context, $m$ denotes the decomposition level, $\psi$ denotes the wavelet type, $\bm{X}_{A_i} \in \mathbb{R}^{C \times L_i}$ and $\bm{X}_{D_i} \in \mathbb{R}^{C \times L_i}$ represent the approximation and detail coefficient series at the $i$-th level of decomposition, respectively. Here, $L_i$ indicates the number of wavelet coefficients in the coefficient series at the $i$-th decomposition level. To avoid information redundancy, we retain only the approximation coefficient series from the final level $m$ while discarding those from levels $1$ through $( m-1 )$, as they are further decomposed into new approximation and detail coefficient series. However, we include the detail coefficient series from all levels in our analysis. In our experiments, we optimized the wavelet type by considering the Daubechies, Symlets, Coiflets, and Biorthogonal wavelet families. 

Each series of wavelet coefficient is processed through a distinct resolution branch within the model, encompassing a RevIN normalization module, a patching and embedding module, multiple mixer modules, a head module, and a RevIN denormalization module. The total number of multivariate coefficient series or resolution branches in the model is given by $(m+1)$ due to the $m$ detail and $1$ approximation coefficient series.

To simplify the notation, we will refer both the approximation coefficient series $\bm{X}_{A_i}$ and the detail coefficient series $\bm{X}_{D_i}$ with $\bm{X}_{W_i} \in \mathbb{R}^{C\times L_i}$ in the following steps. 

\subsubsection{Patching and Embedding module:}
To capture the local information efficiently, we adopt patching and embedding techniques from \cite{c:patchtst}. Each normalized univariate wavelet coefficient series $\underline{\bm{X}}_{W_i}^{(j)} \in \mathbb{R}^{1\times L_i},~j=1,\dots,C,$
is divided into overlapping patches of length $P$. The non-overlapping portion is denoted as stride $S$. Before patching, $\underline{\bm{X}}_{W_i}^{(j)}$ is padded with $S$ number of repeated last values of the sequence $\underline{\bm{X}}_{W_i}^{(j)} $. So, each univariate wavelet coefficient series $\underline{\bm{X}}_{W_i}^{(j)} $ is converted to $\bm{X}_{P_i}^{(j)} \in \mathbb{R}^{1\times N_{i} \times P}$, where $N_i=\frac{(L_i - P)}{S} + 2$ is the number of patches.

The multivariate output of the patching block,
\begin{equation}
    \bm{X}_{P_i} = Patch(\underline{\bm{X}}_{W_i} ) \in \mathbb{R}^{C\times N_{i}\times P}
    \label{eq:patch}
\end{equation}
is passed through a linear embedding layer to encode into $d$ dimensions. This embedding layer is shareable across all variates of $\bm{X}_{P_i}$, i.e.,  
\begin{equation}
    \bm{X}_{d_i} = Embedding(\bm{X}_{P_i}) \in \mathbb{R}^{C\times N_i\times d}.
    \label{eq:embed}
\end{equation}

\subsubsection{Mixer module:}
The Mixer module consists of two primary components, the Patch Mixer and a subsequent Embedding Mixer.  The Patch Mixer functions similarly to the token-mixing MLP as outlined in 
\cite{c:mlpmixervision}. 

Before intermixing information across the patch dimension, 2D-Batch normalization followed by dimension permutation operation is applied on $\bm{X}_{d_i}\in \mathbb{R}^{C\times N_i\times d}$. 
Within the patch mixer, two linear layers are employed alongside the GELU activation function. The first layer expands the dimensionality with factor $t_f$ while the subsequent layer restores it to its original dimension. The operations in the patch mixer can be summarized as,
\begin{equation}
    \bm{X}^{'}_{d_i} = \mathcal{P}(BN(\bm{X}_{d_i})) \in \mathbb{R}^{d\times C\times N_i}
    \label{eq:mix1}
\end{equation}
\begin{equation}
    \bm{X}^{''}_{d_i} = \mathcal{L}_2(\mathcal{G}(\mathcal{L}_1( \bm{X}^{'}_{d_i}))) \in \mathbb{R}^{d\times C\times N_i}
    \label{eq:mix2}
\end{equation}
where $BN(.)$ represents the 2D-Batch normalization, $\mathcal{P(.)}$ represents dimension permutation, $\mathcal{G(.)}$ represents GELU activation, $\mathcal{L}_1: \mathbb{R}^{d\times C\times N_i} \rightarrow \mathbb{R}^{d\times C\times N_i.t_f}$ represents layer-1 and $\mathcal{L}_2: \mathbb{R}^{d\times C\times N_i.t_f} \rightarrow \mathbb{R}^{d\times C\times N_i}$ represents layer-2 in the patch mixer MLP. 

Prior to processing in the Embedding Mixer, $\bm{X}^{''}_{d_i}$ is subjected to dimension permutation and 2D-Batch normalization. In the Embedding Mixer,  $\underline{\bm{X}^{''}_{d_i}}$ traverses two linear layers incorporating GELU activation similarly to the Patch Mixer. However, the initial layer increases the embedding dimensionality $d$ with factor $d_f$, while the subsequent layer restores it to its original dimension. Different than Patch Mixer, a residual connection is also included with the MLP. The operations in the Embedding Mixer can be summarized as,
\begin{equation}
    \underline{\bm{X}^{''}_{d_i}} = BN(\mathcal{P}(\bm{X}^{''}_{d_i}))\in \mathbb{R}^{C\times N_i\times d}
    \label{eq:mix3}
\end{equation}
\begin{equation}
    \bm{X}_{d_{i2}} = \underline{\bm{X}^{''}_{d_i}} + \mathcal{L^{'}}_2(\mathcal{G}(\mathcal{L^{'}}_1(\underline{\bm{X}^{''}_{d_i}})))\in \mathbb{R}^{C\times N_i\times d},
    \label{eq:mix4}
\end{equation}
where $\mathcal{L^{'}}_1: \mathbb{R}^{C\times N_i\times d} \rightarrow \mathbb{R}^{C\times N_i\times d.d_f}$ represents layer-1 and $\mathcal{L^{'}}_2: \mathbb{R}^{C\times N_i\times d.d_f} \rightarrow \mathbb{R}^{C\times N_i\times d}$ represents layer-2.
Two sequential Mixer modules are employed in our model, where the second Mixer module has a residual connection followed by 2D-Batch normalization.

\subsubsection{Head module:}
The Head module comprises a flatten and a linear projection layers. The flatten layer flattens the last two dimensions of the input $Y_{d_i} \in \mathbb{R}^{C \times N_i \times d}$.
\begin{equation}
    \bm{Y}_{f_i} = Flatten(\bm{Y}_{d_i})\in \mathbb{R}^{C \times N_{i}.d},
    \label{eq:flatten}
\end{equation}
and the linear layer transforms $\bm{Y}_{f_i}$ to 
\begin{equation}
    \bm{Y}_{h_i} = Linear(\bm{Y}_{f_i})\in \mathbb{R}^{C \times T_i},
    \label{eq:linearHead}
\end{equation}
where $T_i$ is the prediction length of the wavelet coefficient series.
To determine the value of $T_i$, an auxiliary time series of equivalent length to the predicted series $\bm{X}_T$ undergoes the decomposition module while initializing the model. $T_i$ is set as the length of the auxiliary decomposed wavelet coefficient series.
\subsubsection{Reconstruction:}
The Reconstruction module can be described as,
\begin{equation}
    \bm{Y} = Reconstruction_{\psi}(\bm{Y}_{A_m}, \bm{Y}_{D_m}, \bm{Y}_{D_{m-1}},\ldots, \bm{Y}_{D_1});
    \label{eq:reconstruction}
\end{equation}
where $\bm{Y}_{A_i} \in \mathbb{R}^{C \times T_i}$ and $\bm{Y}_{D_i} \in \mathbb{R}^{C \times T_i}$ are the predicted approximation and detail wavelet coefficient series. $\bm{Y} \in \mathbb{R}^{C \times T}$ is the reconstructed time series, which is transformed by instance denormalization to obtain the final prediction $\bm{X}_T \in \mathbb{R}^{T \times C}$.

\subsubsection{Training:}
$SmoothL1Loss$ is employed to train our model with the default threshold value. Separate dropout values are used for the Embedding and Mixer modules. We used Optuna \cite{c:optuna} with the default setting of Tree-structured Parzen Estimator (TPE) for optimizing the hyperparameters. The optimized hyperparameter values are shown in Table \ref{tab:hyperparameters_list} in Supplementary.
\subsection{Differences with the existing models}
TimeMixer leverages moving average-based seasonal and trend decomposition of multi-scaled time series data and integrates data across multiple scales. WPMixer, on the other hand, employs multi-level wavelet transform-based decomposition, processing each coefficient series individually through a resolution branch. TSMixer incorporates time mixing and channel mixing while WPMixer employs patch mixing followed by embedding mixing. Both TimeMixer and TSMixer handle solely time-domain data, whereas WPMixer extracts features from both the time and frequency domains. Fedformer enhances time series using multi-wavelet transform, frequently converting data between the time and frequency domains. SWformer uses single-level wavelet transform for time series decomposition. However, WPMixer utilizes multi-level wavelet transform, which is computationally less expensive than multi-wavelet transform and more effective than single-level wavelet transform \cite{c:wavelettransform}. Additionally, WPMixer performs time series decomposition at the beginning of the model and reconstructs the series from the predicted coefficient series at the end, avoiding multiple conversions between the time and frequency domains.
\section{Experiments}
\label{Experiments}
We extensively evaluate the long-term forecasting performance of WPMixer on 7 popular datasets: ETTh1, ETTh2, ETTm1, ETTm2, Weather, Electricity, and Traffic. The specifications of datasets are given in Table \ref{tab:dataset_details}. 

\begin{table}[htbp!]
    \centering
    \fontsize{9}{6}\selectfont
    \begin{tabularx}{\columnwidth}{@{}c|c|c|c@{}}
    \toprule
        Dataset & Variates & Dataset Size & Freq. \\ \midrule
        ETTh1, ETTh2 & 7 & (8545, 2881, 2881) & Hourly \\
        ETTm1, ETTm2 & 7 & (34465, 11521, 11521) & 15 min \\
        Weather & 21 & (36792, 5271, 10540) & 10 min \\
        Electricity & 321 & (18317, 2633, 5261) & Hourly \\
        Traffic & 862 & (12185, 1757, 3509) & Hourly \\ \bottomrule
    \end{tabularx}
    \caption{Specifications of the datasets. Dataset size refers to the training, validation, and testing dataset sizes.}
    \label{tab:dataset_details}
\end{table}

\begin{table*}[!htb]
    \fontsize{9}{8}\selectfont
    \begin{tabularx}{\textwidth}{@{}cccccccccccccccccc@{}}
        \multicolumn{2}{c}{\multirow{2}{*}{Models}} & \multicolumn{2}{c}{WPMixer} & \multicolumn{2}{c}{TimeMixer*} & \multicolumn{2}{c}{PatchTST} & \multicolumn{2}{c}{TSMixer} & \multicolumn{2}{c}{TimesNet} & \multicolumn{2}{c}{Crossformer*} & \multicolumn{2}{c}{FiLM*} & \multicolumn{2}{c}{Dlinear*} \\
        \multicolumn{2}{c}{} & \multicolumn{2}{c}{(Ours)} & \multicolumn{2}{c}{2024} & \multicolumn{2}{c}{2023} & \multicolumn{2}{c}{2023} & \multicolumn{2}{c}{2023} & \multicolumn{2}{c}{2023} & \multicolumn{2}{c}{2022a} & \multicolumn{2}{c}{2023} \\
        \toprule
        \multicolumn{2}{c}{Metric} & MSE & MAE & MSE & MAE & MSE & MAE & MSE & MAE & MSE & MAE & MSE & MAE & MSE & MAE & MSE & MAE \\
        \midrule
        \multirow{5}{*}{ETTh1} & 96 & \textbf{0.347} & \textbf{0.383} & \underline{0.361} & \underline{0.390} & 0.370 & 0.400 & \underline{0.361} & 0.392 & 0.384 & 0.402 & 0.418 & 0.438 & 0.422 & 0.432 & 0.375 & 0.399 \\
         & 192 & \textbf{0.381} & \textbf{0.408} & 0.409 & \underline{0.414} & 0.413 & 0.429 & \underline{0.404} & 0.418 & 0.436 & 0.429 & 0.539 & 0.517 & 0.462 & 0.458 & 0.405 & 0.416 \\
         & 336 & \textbf{0.382} & \textbf{0.412} & 0.430 & \underline{0.429} & 0.422 & 0.440 & \underline{0.420} & 0.431 & 0.491 & 0.469 & 0.709 & 0.638 & 0.501 & 0.483 & 0.439 & 0.443 \\
         & 720 & \textbf{0.405} & \textbf{0.432} & \underline{0.445} & \underline{0.460} & 0.447 & 0.468 & 0.463 & 0.472 & 0.521 & 0.500 & 0.733 & 0.636 & 0.544 & 0.526 & 0.472 & 0.490 \\
         \cmidrule{2-18}
         & Avg & \textbf{0.379} & \textbf{0.409} & \underline{0.411} & \underline{0.423} & 0.413 & 0.434 & 0.412 & 0.428 & 0.458 & 0.450 & 0.600 & 0.557 & 0.482 & 0.475 & 0.423 & 0.437 \\
         \midrule
        \multirow{5}{*}{ETTh2} & 96 & \textbf{0.253} & \textbf{0.328} & \underline{0.271} & \underline{0.330} & 0.274 & 0.337 & 0.274 & 0.341 & 0.340 & 0.374 & 0.425 & 0.463 & 0.323 & 0.370 & 0.289 & 0.353 \\
         & 192 & \textbf{0.303} & \textbf{0.364} & \underline{0.317} & 0.402 & 0.341 & \underline{0.382} & 0.339 & 0.385 & 0.402 & 0.414 & 0.473 & 0.500 & 0.391 & 0.415 & 0.383 & 0.418 \\
         & 336 & \textbf{0.305} & \textbf{0.371} & 0.332 & 0.396 & \underline{0.329} & \underline{0.384} & 0.361 & 0.406 & 0.452 & 0.452 & 0.581 & 0.562 & 0.415 & 0.440 & 0.448 & 0.465 \\
         & 720 & \underline{0.373} & \underline{0.417} & \textbf{0.342} & \textbf{0.408} & 0.379 & 0.422 & 0.445 & 0.470 & 0.462 & 0.468 & 0.775 & 0.665 & 0.441 & 0.459 & 0.605 & 0.551 \\
          \cmidrule{2-18}
         & Avg & \textbf{0.309} & \textbf{0.370} & \underline{0.316} & 0.384 & 0.331 & \underline{0.381} & 0.355 & 0.401 & 0.414 & 0.427 & 0.564 & 0.548 & 0.393 & 0.421 & 0.431 & 0.447 \\
         \midrule
        \multirow{5}{*}{ETTm1} & 96 & \textbf{0.275} & \textbf{0.333} & 0.291 & 0.340 & 0.293 & 0.346 & \underline{0.285} & \underline{0.339} & 0.338 & 0.375 & 0.361 & 0.403 & 0.302 & 0.345 & 0.299 & 0.343 \\
         & 192 & \textbf{0.319} & \textbf{0.362} & \underline{0.327} & \underline{0.365} & 0.333 & 0.370 & \underline{0.327} & \underline{0.365} & 0.374 & 0.387 & 0.387 & 0.422 & 0.338 & 0.368 & 0.335 & \underline{0.365} \\
         & 336 & \textbf{0.347} & 0.384 & 0.360 & \textbf{0.381} & 0.369 & 0.392 & \underline{0.356} & \underline{0.382} & 0.410 & 0.411 & 0.605 & 0.572 & 0.373 & 0.388 & 0.369 & 0.386 \\
         & 720 & \textbf{0.403} & \textbf{0.414} & \underline{0.415} & 0.417 & 0.416 & 0.420 & 0.419 & \textbf{0.414} & 0.478 & 0.450 & 0.703 & 0.645 & 0.420 & 0.420 & 0.425 & 0.421 \\
          \cmidrule{2-18}
         & Avg & \textbf{0.336} & \textbf{0.373} & 0.348 & \underline{0.375} & 0.353 & 0.382 & \underline{0.347} & \underline{0.375} & 0.400 & 0.406 & 0.514 & 0.510 & 0.358 & 0.380 & 0.357 & 0.379 \\
         \midrule
        \multirow{5}{*}{ETTm2} & 96 & \textbf{0.159} & \textbf{0.246} & 0.164 & 0.254 & 0.166 & 0.256 & \underline{0.163} & \underline{0.252} & 0.187 & 0.267 & 0.275 & 0.358 & 0.165 & 0.256 & 0.167 & 0.260 \\
         & 192 & \textbf{0.214} & \textbf{0.286} & 0.223 & 0.295 & 0.223 & 0.296 & \underline{0.216} & \underline{0.290} & 0.249 & 0.309 & 0.345 & 0.400 & 0.222 & 0.296 & 0.224 & 0.303 \\
         & 336 & \textbf{0.266} & \textbf{0.322} & 0.279 & 0.330 & 0.274 & 0.329 & \underline{0.268} & \underline{0.324} & 0.321 & 0.351 & 0.657 & 0.528 & 0.277 & 0.333 & 0.281 & 0.342 \\
         & 720 & \textbf{0.344} & \textbf{0.374} & \underline{0.359} & \underline{0.383} & 0.362 & 0.385 & 0.420 & 0.422 & 0.408 & 0.403 & 1.208 & 0.753 & 0.371 & 0.389 & 0.397 & 0.421 \\
          \cmidrule{2-18}
         & Avg & \textbf{0.246} & \textbf{0.307} & \underline{.256} & \underline{0.315} & \underline{0.256} & 0.317 & 0.267 & 0.322 & 0.291 & 0.333 & 0.621 & 0.510 & 0.259 & 0.319 & 0.267 & 0.332 \\
         \midrule
        \multirow{5}{*}{Weather} & 96 & \textbf{0.141} & \textbf{0.188} & 0.147 & \underline{0.197} & 0.149 & 0.198 & \underline{0.145} & 0.198 & 0.172 & 0.220 & 0.232 & 0.302 & 0.199 & 0.262 & 0.176 & 0.237 \\
         & 192 & \textbf{0.185} & \textbf{0.229} & \underline{0.189} & \underline{0.239} & 0.194 & 0.241 & 0.191 & 0.242 & 0.219 & 0.261 & 0.371 & 0.410 & 0.228 & 0.288 & 0.220 & 0.282 \\
         & 336 & \textbf{0.236} & \textbf{0.271} & \underline{0.241} & \underline{0.280} & 0.245 & 0.282 & 0.242 & \underline{0.280} & 0.280 & 0.306 & 0.495 & 0.515 & 0.267 & 0.323 & 0.265 & 0.319 \\
         & 720 & \textbf{0.307} & \textbf{0.321} & \underline{0.310} & \underline{0.330} & 0.314 & 0.334 & 0.320 & 0.336 & 0.365 & 0.359 & 0.526 & 0.542 & 0.319 & 0.361 & 0.323 & 0.362 \\
          \cmidrule{2-18}
         & Avg & \textbf{0.217} & \textbf{0.252} & \underline{0.222} & \underline{0.262} & 0.226 & 0.264 & 0.225 & 0.264 & 0.259 & 0.287 & 0.406 & 0.442 & 0.253 & 0.309 & 0.246 & 0.300 \\
         \midrule
        \multirow{5}{*}{Electricity} & 96 & \textbf{0.128} & \textbf{0.222} & \underline{0.129} & 0.224 & \underline{0.129} & \textbf{0.222} & 0.131 & 0.229 & 0.168 & 0.272 & 0.150 & 0.251 & 0.154 & 0.267 & 0.140 & 0.237 \\
         & 192 & \underline{0.145} & \underline{0.237} & \textbf{0.140} & \textbf{0.220} & 0.147 & 0.240 & 0.151 & 0.246 & 0.184 & 0.289 & 0.161 & 0.260 & 0.164 & 0.258 & 0.153 & 0.249 \\
         & 336 & \textbf{0.161} & \underline{0.256} & \textbf{0.161} & \textbf{0.255} & 0.163 & 0.259 & \textbf{0.161} & 0.261 & 0.198 & 0.300 & 0.182 & 0.281 & 0.188 & 0.283 & 0.169 & 0.267 \\
         & 720 & \underline{0.196} & \textbf{0.287} & \textbf{0.194} & \textbf{0.287} & 0.197 & 0.290 & 0.197 & 0.293 & 0.220 & 0.320 & 0.251 & 0.339 & 0.236 & 0.332 & 0.203 & 0.301 \\
          \cmidrule{2-18}
         & Avg & \underline{0.158} & \underline{0.251} & \textbf{0.156} & \textbf{0.246} & 0.159 & 0.253 & 0.160 & 0.257 & 0.192 & 0.295 & 0.186 & 0.283 & 0.186 & 0.285 & 0.166 & 0.264 \\
         \midrule
        \multirow{5}{*}{Traffic} & 96 & \textbf{0.354} & \textbf{0.246} & \underline{0.360} & \underline{0.249} & \underline{0.360} & \underline{0.249} & 0.376 & 0.264 & 0.593 & 0.321 & 0.514 & 0.267 & 0.416 & 0.294 & 0.410 & 0.282 \\
         & 192 & \textbf{0.371} & 0.253 & \underline{0.375} & \textbf{0.250} & 0.379 & 0.256 & 0.397 & 0.277 & 0.617 & 0.336 & 0.549 & \underline{0.252} & 0.408 & 0.288 & 0.423 & 0.287 \\
         & 336 & \underline{0.387} & \underline{0.267} & \textbf{0.385} & 0.270 & 0.392 & \textbf{0.264} & 0.413 & 0.290 & 0.629 & 0.336 & 0.530 & 0.300 & 0.425 & 0.298 & 0.436 & 0.296 \\
         & 720 & \underline{0.431} & 0.289 & \textbf{0.430} & \textbf{0.281} & 0.432 & \underline{0.286} & 0.444 & 0.306 & 0.640 & 0.350 & 0.573 & 0.313 & 0.520 & 0.353 & 0.466 & 0.315 \\
          \cmidrule{2-18}
         & Avg & \textbf{0.386} & \underline{0.264} & \underline{0.387} & \textbf{0.262} & 0.391 & \underline{0.264} & 0.408 & 0.284 & 0.620 & 0.336 & 0.542 & 0.283 & 0.442 & 0.308 & 0.434 & 0.295 \\
         \midrule
        \multicolumn{1}{l}{} & \multicolumn{1}{l}{1st Count:} & 29 & 26 & 7 & 9 & 0 & 2 & 1 & 1 & 0 & 0 & 0 & 0 & 0 & 0 & 0 & 0 \\
        \bottomrule
    \end{tabularx}
    \caption{Multivariate long-term forecasting results. Four commonly used prediction lengths (96,192,336,720) from the literature are considered for each dataset. The length of the look-back window is a hyperparameter. The results of the models marked with $*$ are taken from \cite{c:timemixer}; other results are taken from the corresponding papers.}
    \label{tab:search_result}
\end{table*}

\subsubsection{Baselines:}
We compare WPMixer with seven recent time series forecasting methods, namely TimeMixer \cite{c:timemixer}), TSMixer \cite{c:tsmixer}, TimesNet \cite{c:timesnet}, FiLM \cite{c:film}, DLinear \cite{c:dlinear}, PatchTST \cite{c:patchtst}, and Crossformer \cite{c:crossformer}. TimeMixer and TSMixer, which can be considered as the state-of-the-art models based on their performances on the benchmark datasets, derive their architectures from the MLP-Mixer model while PatchTST and Crossformer utilize transformer architectures.
\subsubsection{Setup:}
Following the practice in Informer, Autoformer, PatchTST, TSMixer, and TimeMixer, all datasets were normalized to a zero mean and unit standard deviation. The normalized datasets served as the basis for ground truth in our evaluations. In long-term forecasting, the lengths of predictions were set at 96, 192, 336, and 720, in alignment with prior studies. During the training phase, SmoothL1Loss was employed, whereas Mean Squared Error (MSE) and Mean Absolute Error (MAE) were utilized for evaluation purposes. 
Experiments with the ETT and Weather datasets were performed on a single NVIDIA GeForce RTX 4090 GPU while the experiments with the Electricity and Traffic datasets were carried out using two NVIDIA A100 GPUs.
\subsection{Multivariate Long-Term Forecasting Results}
In long-term multivariate time series forecasting, existing studies employed distinct look-back window lengths to optimize performance. For a comprehensive comparison, we present our results under two experimental setups following TimeMixer \cite{c:timemixer}.

In the first setup, we calibrated the look-back window length alongside other hyperparameters to enhance forecasting accuracy. We determined the optimal look-back window lengths for each dataset, exploring values of 96, 192, 336, 512, 1024, and 1200. The comprehensive results under this setup are presented in Table \ref{tab:search_result} while the optimized hyperparameter values and run information are given in Table \ref{tab:hyperparameters_list} in Supplementary. The performance of other models listed in Table \ref{tab:search_result} are also their optimized results \cite{c:timemixer}. Our analysis revealed that our model's performance is notably superior compared to its counterparts. Specifically, our model decreased MSE on average across the ETTh1, ETTh2, ETTm1, and ETTm2 datasets by 7.8\%, 2.2\%, 3.4\%, and 3.9\%, respectively. Similarly, MAE was reduced by 3.3\%, 6.4\%, 0.5\%, and 2.5\%, respectively, for these datasets. On the Weather and Traffic datasets, our model demonstrated lower MSE and MAE in average prediction relative to the state-of-the-art TimeMixer model. Moreover, on the Electricity dataset, our model achieved the highest performance following the TimeMixer model.

In the second setup, we followed the unified setting of TimeMixer for all the datasets. The detailed results are presented in Table \ref{tab:unified_result} in Supplementary. We achieved lower MSE and MAE scores on average on the ETT and Electricity datasets compared to the TimeMixer model. 

\subsection{Computational efficiency and robustness} 
We evaluate WPMixer's computational cost in terms of the number of giga floating point operations (GFLOPs), a hardware-independent metric. We compute the GFLOPs for WPMixer and TimeMixer using the unified setting outlined by \cite{c:timemixer} with embedding dimension $d = 16$ for the ETTh1 dataset. The comparison is presented in Table \ref{tab:table_gflops}. WPMixer consistently requires less than one tenth GFLOPs across all prediction lengths compared to TimeMixer.

We also evaluate our model with three different random seeds by computing the mean and standard deviation for MSE and MAE. Results are averaged over the prediction lengths of 96, 192, 336, and 720. As shown in Table \ref{tab:robustness}, our model exhibits a lower standard deviation than TimeMixer in all cases, highlighting the robustness of our approach.
\begin{table}[htbp!]
    \centering
    \fontsize{9}{7}\selectfont
    \begin{tabularx}{\columnwidth}{@{}c|c|ccc|ccc@{}}
        \toprule
         &  & \multicolumn{3}{c|}{WPMixer} & \multicolumn{3}{c}{TimeMixer} \\ \midrule
         & T & MSE & MAE & GFLOPs & MSE & MAE & GFLOPs \\ \midrule
        \multirow{4}{*}{\begin{sideways}ETTh1\end{sideways}} & 96 & 0.370 & 0.390 & 0.210 & 0.375 & 0.400 & 2.774 \\
         & 192 & 0.424 & 0.420 & 0.226 & 0.429 & 0.421 & 3.281 \\
         & 336 & 0.462 & 0.433 & 0.211 & 0.484 & 0.458 & 4.040 \\
         & 720 & 0.455 & 0.449 & 0.481 & 0.498 & 0.482 & 6.066 \\ \bottomrule
    \end{tabularx}
    
    \caption{WPMixer is ten folds more efficient for $d=16$.}
    \label{tab:table_gflops}
\end{table}
\begin{table}[!htb]
    \fontsize{9}{9}\selectfont
    \begin{tabularx}{\columnwidth}{@{}c|cc|cc@{}}
        \toprule
        \multirow{2}{*}{} & \multicolumn{2}{c|}{WPMixer} & \multicolumn{2}{c}{TimeMixer} \\
         & MSE & MAE & MSE & MAE \\ \midrule
        (1) & 0.422 ± 0.001 & 0.423 ± 0.001 & 0.447 ± 0.002 & 0.440 ± 0.005 \\
        (2) & 0.355 ± 0.003 & 0.387 ± 0.001 & 0.364 ± 0.008 & 0.395 ± 0.010 \\
        (3) & 0.376 ± 0.002 & 0.388 ± 0.001 & 0.381 ± 0.003 & 0.395 ± 0.006 \\
        (4) & 0.271 ± 0.001 & 0.317 ± 0.001 & 0.275 ± 0.001 & 0.323 ± 0.003 \\
        (5) & 0.243 ± 0.001 & 0.269 ± 0.000 & 0.240 ± 0.010 & 0.271 ± 0.009 \\
        (6) & 0.177 ± 0.000 & 0.267 ± 0.000 & 0.182 ± 0.017 & 0.272 ± 0.006 \\
        (7) & 0.489 ± 0.005 & 0.297 ± 0.001 & 0.484 ± 0.015 & 0.297 ± 0.013 \\ \bottomrule
    \end{tabularx}
    \caption{Model robustness under the unified setting, including similar look-back window length, batch size, and epochs for all models. (1), (2), (3), (4), (5), (6), and (7) refer to ETTh1, ETTh2, ETTm1, ETTm2, Weather, Electricity, and Traffic datasets, respectively.}
    \label{tab:robustness}
\end{table}
\begin{table}[!htb]
    \centering
    \fontsize{9}{6}\selectfont
    \begin{tabularx}{\columnwidth}{c|cccccc|c|c|c|c}
        \toprule
         & \multicolumn{6}{c|}{Modules} & \begin{sideways}ETTh1\end{sideways} & \begin{sideways}ETTh2\end{sideways} & \begin{sideways}ETTm1\end{sideways} & \begin{sideways}ETTm2\end{sideways} \\ \midrule
        Case & $D$ & $P$ & $E$ & $P_x$ & $E_x$ & $H$ & MSE & MSE & MSE & MSE \\ \midrule
        I & \checkmark & \checkmark & \checkmark & \checkmark & \checkmark & \checkmark & 0.379 & \textbf{0.308} & \textbf{0.336} & \textbf{0.245} \\
        II & $\times$ & \checkmark & \checkmark & \checkmark & \checkmark & \checkmark & 0.388 & 0.311 & 0.339 & 0.247 \\
        III & \checkmark & $\times$ & $\times$ & \checkmark & \checkmark & \checkmark & 0.384 & 0.316 & 0.339 & 0.250 \\
        IV & $\times$ & $\times$ & $\times$ & \checkmark & \checkmark & \checkmark & 0.392 & 0.325 & 0.345 & 0.249 \\
        V & \checkmark & \checkmark & $\times$ & \checkmark & \checkmark & \checkmark & 0.378 & 0.314 & 0.339 & 0.247 \\
        VI & $\times$ & \checkmark & $\times$ & \checkmark & \checkmark & \checkmark & 0.390 & 0.320 & 0.343 & 0.248 \\
        VII & \checkmark & \checkmark & \checkmark & $\times$ & $\times$ & \checkmark & 0.394 & 0.311 & 0.353 & 0.252 \\
        VIII & $\times$ & \checkmark & \checkmark & $\times$ & $\times$ & \checkmark & 0.399 & 0.312 & 0.354 & 0.252 \\
        IX & \checkmark & $\times$ & $\times$ & $\times$ & $\times$ & \checkmark & 0.400 & 0.315 & 0.356 & 0.251 \\
        X & $\times$ & $\times$ & $\times$ & $\times$ & $\times$ & \checkmark & 0.403 & 0.315 & 0.355 & 0.252 \\
        XI & \checkmark & \checkmark & $\times$ & $\times$ & $\times$ & \checkmark & 0.400 & 0.312 & 0.355 & 0.251 \\
        XII & $\times$ & \checkmark & $\times$ & $\times$ & $\times$ & \checkmark & 0.403 & 0.314 & 0.355 & 0.252 \\
        XIII & \checkmark & \checkmark & \checkmark & $\times$ & \checkmark & \checkmark & \textbf{0.377} & 0.314 & 0.339 & 0.247 \\
        XIV & $\times$ & \checkmark & \checkmark & $\times$ & \checkmark & \checkmark & 0.392 & 0.314 & 0.342 & 0.248 \\ \bottomrule
    \end{tabularx}
    \caption{Contribution of each module in WPMixer. $D$, $P$, $E$, $P_x$, $E_x$, and $H$ refer to the decomposition, patch, embedding, patch mixer, embedding mixer, and head modules, respectively. Look-back window is set to 512. Results are averaged over the prediction lengths 96, 192, 336, and 720.}
    \label{tab:ablation_result_short}
\end{table}
\subsection{Ablation Study}
\subsubsection{WPMixer modules:}
We conducted an extensive ablation study to evaluate the individual contribution of each module within the proposed model using the ETT datasets. This analysis consists of fourteen distinct cases, each exploring a different combination of the modules. Case-$I$ represents the foundational architecture of WPMixer. The details of the other cases are delineated in Table \ref{tab:ablation_result_short}. For each case, we performed a thorough search of optimum hyperparameters utilizing Optuna. The results in Table \ref{tab:ablation_result_short} demonstrate the importance of all proposed modules.
\begin{figure}[!htb]
    \centering
    \includegraphics[width=.8\columnwidth]{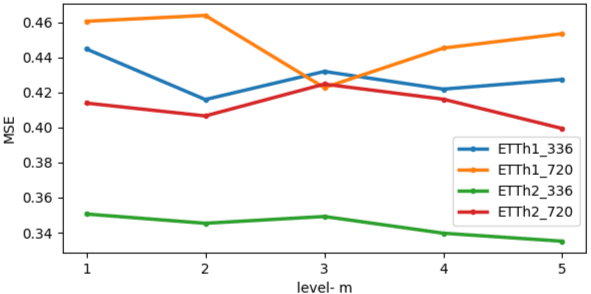}
    \caption{WPMixer performance with the varying level of the decomposition $m$.}
    \label{fig:level_vs_mse}
\end{figure}
\subsubsection{Effect of multiple levels of decomposition:}
We assessed the impact of multi-level decomposition by varying $m$ from 1 to 5. The other parameters are kept fixed for all $m$ as follows, look-back window $512$, initial learning rate $0.001$, wavelet type Daubechies 5, batch size $128$, epochs $10$, $d=256$, $t_f=7$, $d_f=7$, patch size $16$, and stride $8$. MSE performances for prediction lengths of 336 and 720 on the ETTh datasets are presented in Figure \ref{fig:level_vs_mse}. The results indicate that the optimal level $m$ depends on the prediction length and dataset. Consequently, we treated $m$ as a hyperparameter in our model and performed a search to identify its optimal value for every experiment.

\subsubsection{SmoothL1 vs MSE loss:}
In our experiments, we utilized the $SmoothL1$ loss as the primary loss function instead of the traditional $MSE$ loss. We conducted an ablation study using the ETTh2 and ETTm2 datasets, employing an exhaustive search across the hyperparameter space. Detailed findings are presented in Table \ref{tab:mse_vs_smoothL1}. Analysis of the results from Table \ref{tab:mse_vs_smoothL1} demonstrates that the adoption of the $SmoothL1$ loss improves the performance of our model.
\begin{figure}[h]
    \centering
    \begin{subfigure}{0.49\columnwidth}
        \centering
        \includegraphics[width=\textwidth]{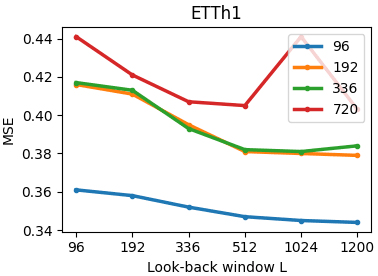}
        \caption{ETTh1}
    \end{subfigure}
    \hfill
    \begin{subfigure}{0.49\columnwidth}
        \centering
        \includegraphics[width=\textwidth]{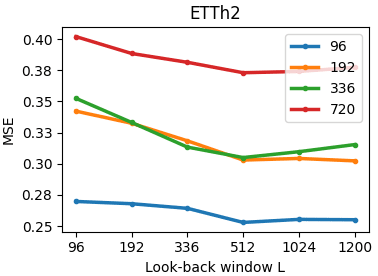}
        \caption{ETTh2}
    \end{subfigure}
    \caption{Performance of the model with increasing look-back window length $L$.}
    \label{fig:look_back_winodw}
\end{figure}

\begin{table}[!htbp]
    \centering
    \fontsize{9}{6}\selectfont
    \begin{tabularx}{\columnwidth}{c|cccc|cccc}
        \toprule
         & \multicolumn{4}{c|}{ETTm2} & \multicolumn{4}{c}{ETTh2} \\ \midrule
         & \multicolumn{2}{c}{$MSEloss$} & \multicolumn{2}{c|}{$SmoothL1$} & \multicolumn{2}{c}{$MSEloss$} & \multicolumn{2}{c}{$SmoothL1$} \\ \midrule
        T & MSE & MAE & MSE & MAE & MSE & MAE & MSE & MAE \\ \midrule
        96 & 0.165 & 0.257 & 0.159 & 0.246 & 0.251 & 0.327 & 0.253 & 0.328 \\
        192 & 0.219 & 0.291 & 0.214 & 0.286 & 0.308 & 0.365 & 0.303 & 0.364 \\
        336 & 0.271 & 0.327 & 0.266 & 0.322 & 0.306 & 0.373 & 0.305 & 0.371 \\
        720 & 0.349 & 0.384 & 0.344 & 0.374 & 0.374 & 0.419 & 0.373 & 0.417 \\ \bottomrule
    \end{tabularx}
    \caption{$SmoothL1$ loss vs. $MSE$ loss for training.}
    \label{tab:mse_vs_smoothL1}
\end{table}
\subsubsection{Look-back window:}
We also evaluated the impact of look-back window size on the forecasting performance using the ETTh datasets, as illustrated in Figure \ref{fig:look_back_winodw}. While in general the MSE value is reduced with increasing look-back window length, after a certain length, the model's performance stops improving or even degrades in some cases such as the prediction length of 336.

\section{Conclusion}
\label{Conclusion}
In this study, we introduced the Wavelet Patch Mixer (WPMixer), a computationally efficient long-term time series forecasting model. Our model utilizes multi-level wavelet decomposition to capture multi-resolution information in both the time and frequency domains. By incorporating patching for local information and a patch mixer for global information, we enhanced the model's capability to handle complex characteristics and abrupt spikes and dips in real-world data. The addition of an embedding mixer after each patch mixer further improved the model's forecasting performance. Our experimental results demonstrated that WPMixer achieves state-of-the-art performance efficiently in various long-term forecasting tasks. Through comprehensive experiments, we analyzed the model performance, computational cost, robustness to random initializations, effects of decomposition level, loss function, and look-back window size.

\section*{Acknowledgements}
This work was supported by the U.S. National Institute of Food and Agriculture under Grant 2023-67019-38829.

\bibliography{aaai25}


\clearpage
\onecolumn
\section{Supplementary for WPMixer: Efficient Multi-Resolution Mixing for Long-Term Time Series Forecasting}

\subsection{Hyperparameter Tuning}
The results in Table \ref{tab:search_result} in the main paper are obtained from a single run with random seed 42 and the hyperparameter values given in Table \ref{tab:hyperparameters_list}. The hyperparameter values explored during the hyperparameter tuning are presented in Table \ref{tab:hyper_parameters_range}.

\begin{table*}[!htbp]
    \fontsize{9}{6}\selectfont
    \centering
    \begin{tabularx}{\textwidth}{@{}c|c|c|c|c|c|c|c|c|c|c|c|c|c|c@{}}
        \toprule
         & Pred len & Look back & Initial lr & Batch & $\psi$ & $m$ & $t_f$ & $d_f$ & Mixer d/o & Embed d/o & Patch & Stride & $d$ & Epochs \\ \midrule
        \multirow{4}{*}{ETTh1} & 96 & 512 & 0.00024 & 256 & db2 & 2 & 5 & 8 & 0.4 & 0.1 & 16 & 8 & 256 & 30 \\
         & 192 & 512 & 0.0002 & 256 & db3 & 2 & 5 & 5 & 0.05 & 0.2 & 16 & 8 & 256 & 30 \\
         & 336 & 512 & 0.00013 & 256 & db2 & 1 & 3 & 3 & 0 & 0.4 & 16 & 8 & 256 & 30 \\
         & 720 & 512 & 0.00024 & 256 & db2 & 1 & 5 & 3 & 0.2 & 0.4 & 16 & 8 & 128 & 30 \\  \midrule
        \multirow{4}{*}{ETTh2} & 96 & 512 & 0.00047 & 256 & db2 & 2 & 5 & 5 & 0 & 0.1 & 16 & 8 & 256 & 30 \\
         & 192 & 512 & 0.00029 & 256 & db2 & 3 & 3 & 8 & 0 & 0 & 16 & 8 & 256 & 30 \\
         & 336 & 512 & 0.00062 & 256 & db2 & 5 & 5 & 3 & 0.1 & 0.1 & 16 & 8 & 128 & 30 \\
         & 720 & 512 & 0.00081 & 256 & db2 & 5 & 5 & 5 & 0.4 & 0 & 16 & 8 & 128 & 30 \\  \midrule
        \multirow{4}{*}{ETTm1} & 96 & 512 & 0.00128 & 256 & db2 & 1 & 5 & 3 & 0.4 & 0.2 & 48 & 24 & 256 & 80 \\
         & 192 & 512 & 0.00242 & 256 & db3 & 1 & 3 & 7 & 0.4 & 0.05 & 48 & 24 & 128 & 80 \\
         & 336 & 512 & 0.00159 & 256 & db5 & 1 & 7 & 7 & 0.4 & 0 & 48 & 24 & 256 & 80 \\
         & 720 & 512 & 0.00201 & 256 & db5 & 4 & 3 & 8 & 0.4 & 0.05 & 48 & 24 & 128 & 80 \\  \midrule
        \multirow{4}{*}{ETTm2} & 96 & 512 & 0.00077 & 256 & bior3.1 & 1 & 3 & 8 & 0.4 & 0 & 48 & 24 & 256 & 80 \\
         & 192 & 512 & 0.00028 & 256 & db2 & 1 & 3 & 7 & 0.2 & 0.1 & 48 & 24 & 256 & 80 \\
         & 336 & 512 & 0.00023 & 256 & db2 & 1 & 3 & 5 & 0.4 & 0 & 48 & 24 & 256 & 80 \\
         & 720 & 512 & 0.00104 & 256 & db2 & 1 & 3 & 8 & 0.4 & 0 & 48 & 24 & 256 & 80 \\  \midrule
        \multirow{4}{*}{Weather} & 96 & 512 & 0.00091 & 32 & db3 & 2 & 3 & 7 & 0.4 & 0.1 & 16 & 8 & 256 & 60 \\
         & 192 & 512 & 0.00138 & 64 & db3 & 1 & 3 & 7 & 0.4 & 0 & 16 & 8 & 128 & 60 \\
         & 336 & 512 & 0.00061 & 32 & db3 & 2 & 7 & 7 & 0.4 & 0.4 & 16 & 8 & 128 & 60 \\
         & 720 & 512 & 0.00223 & 128 & db2 & 3 & 7 & 5 & 0.1 & 0.4 & 16 & 8 & 256 & 60 \\  \midrule
        \multirow{4}{*}{Electricity} & 96 & 512 & 0.00328 & 32 & sym3 & 2 & 3 & 5 & 0.1 & 0 & 16 & 8 & 32 & 100 \\
         & 192 & 512 & 0.00049 & 32 & coif5 & 3 & 7 & 5 & 0.1 & 0.05 & 16 & 8 & 32 & 100 \\
         & 336 & 512 & 0.00251 & 32 & sym4 & 1 & 5 & 7 & 0.2 & 0.05 & 16 & 8 & 32 & 100 \\
         & 720 & 512 & 0.00198 & 32 & db2 & 2 & 7 & 8 & 0.1 & 0 & 16 & 8 & 32 & 100 \\  \midrule
        \multirow{4}{*}{Traffic} & 96 & 1200 & 0.00104 & 16 & db3 & 1 & 3 & 5 & 0.05 & 0.05 & 16 & 8 & 16 & 60 \\
         & 192 & 1200 & 0.00057 & 16 & db3 & 1 & 3 & 5 & 0.05 & 0 & 16 & 8 & 32 & 60 \\
         & 336 & 1200 & 0.00103 & 16 & bior3.1 & 1 & 7 & 7 & 0 & 0.1 & 16 & 8 & 32 & 50 \\
         & 720 & 1200 & 0.0015 & 16 & db3 & 1 & 7 & 3 & 0.05 & 0.2 & 16 & 8 & 32 & 60 \\ \bottomrule
    \end{tabularx}
    \caption{Comprehensive hyperparameter tuning for the multi-variate long-term forecasting task, optimized using Optuna. The hyperparameters are described in the main paper in the following places: $\psi$ (wavelet type) and decomposition level $m$ in Eq. \ref{eq:pmd1}; expansion factors $t_f$ and $d_f$ in Eqs. \ref{eq:mix2} and \ref{eq:mix4}, respectively; embedding dimension $d$ in Eq. \ref{eq:embed}. "Mixer d/o" and "Embed d/o" denote the distinct dropouts employed in the Mixer module and the Embedding layer. Learning rate is gradually reduced in the training using the formula, $lr := lr \times 0.9^{(epoch -3)}$.} 
    \label{tab:hyperparameters_list}
\end{table*}

\begin{table*}[!htbp]
    \centering
    \fontsize{9}{7}\selectfont
    \begin{tabular}{@{}c|c|c|c|c|c@{}}
        \toprule
         & ETTh & ETTm & Weather & Electricity & Traffic \\ \midrule
        Batch & 256 & 256 & 32, 64, 128 & 32 & 16 \\ \midrule
        $d$ & 128, 256 & 128, 256 & 128, 256 & 16, 32 & 16, 32 \\ \midrule
        Look back window& \multicolumn{5}{c}{96, 192, 336, 512, 1024, 1200} \\ \midrule
        Initial lr & \multicolumn{5}{c}{max 0.01, min 0.00001} \\ \midrule
        $\psi$ & \multicolumn{5}{c}{db2,   db3, db5, sym2, sym3, sym4, sym5, coif4, coif5, bior3.1, bior3.5} \\ \midrule
        $m$ & \multicolumn{5}{c}{1, 2, 3, 4, 5} \\ \midrule
        $t_f$ & \multicolumn{5}{c}{3, 5, 7, 9} \\ \midrule
        $d_f$ & \multicolumn{5}{c}{3, 5, 7, 8, 9} \\ \midrule
        Mixer Dropout & \multicolumn{5}{c}{0.0, 0.05, 0.1, 0.2, 0.4} \\ \midrule
        Embedding Layer Dropout & \multicolumn{5}{c}{0.0, 0.05, 0.1, 0.2, 0.4} \\ \midrule
        Patch-Stride & \multicolumn{5}{c}{16-8, 32-16, 48-24} \\ \bottomrule
    \end{tabular}
    \caption{Hyperparameters search space. Wavelet type db*, sym*, coif*, and bior* refer to the Daubechies, Symlets, Coiflets, and Biorthogonal wavelet family, respectively.}
    \label{tab:hyper_parameters_range}
\end{table*}

\subsection{Computing Device Configuration}
WPMixer is implemented in PyTorch, version 3.10.12. Experiments with the ETT and Weather datasets are conducted on a single NVIDIA GeForce RTX 4090 GPU (24 GB), while experiments with the Electricity and Traffic datasets are conducted on two NVIDIA A100 GPUs (total 160 GB). The CPU specifications used in the experiments are processor AMD Ryzen 9 7950X 16-core and RAM 128 GB. The operating system is Windows 11.

\subsection{Multivariate Long-Term Forecasting under Unified Setting}
We employ the unified experimental setting of TimeMixer, including the look-back window, batch size, and epochs, for all the datasets while optimizing other hyperparameters, including learning rate, wavelet type $\psi$, $m$, $t_f$, $d_f$, dropouts, patch size $P$, stride $S$, and embedding dimension $d$. 
We ran all experiments three times with three random seed values and averaged the results. The detailed results are presented in Table \ref{tab:unified_result}. Our model reduces the MSE values for the average prediction across ETTh1, ETTh2, ETTm1, ETTm2, and Electricity datasets by 5.6$\%$, 2.5$\%$, 1.3$\%$, 1.5$\%$, and 2.7$\%$ while reducing the MAE values by 3.9$\%$, 2.0$\%$, 1.8$\%$, 1.9$\%$, and 1.8$\%$.

\begin{table*}[!htb]
    \fontsize{9}{6}\selectfont
    \begin{tabularx}{\textwidth}{@{}cccccccccccccccccccc@{}}
        \multicolumn{2}{c}{Models} & \multicolumn{2}{c}{\begin{tabular}[c]{@{}c@{}}WPMixer\\      (Ours)\end{tabular}} & \multicolumn{2}{c}{\begin{tabular}[c]{@{}c@{}}TimeMixer\\      (2024)\end{tabular}} & \multicolumn{2}{c}{\begin{tabular}[c]{@{}c@{}}iTransformer*\\      (2024)\end{tabular}} & \multicolumn{2}{c}{\begin{tabular}[c]{@{}c@{}}TSMixer\\      (2023)\end{tabular}} & \multicolumn{2}{c}{\begin{tabular}[c]{@{}c@{}}PatchTST\\      (2023)\end{tabular}} & \multicolumn{2}{c}{\begin{tabular}[c]{@{}c@{}}TimesNet\\      (2023)\end{tabular}} & \multicolumn{2}{c}{\begin{tabular}[c]{@{}c@{}}Crossformer\\      (2023)\end{tabular}} & \multicolumn{2}{c}{\begin{tabular}[c]{@{}c@{}}FiLM\\      (2022a)\end{tabular}} & \multicolumn{2}{c}{\begin{tabular}[c]{@{}c@{}}Dlinear\\      (2023)\end{tabular}} \\ \midrule
        \multicolumn{2}{c}{Metric} & MSE & MAE & \multicolumn{1}{l}{MSE} & \multicolumn{1}{l}{MAE} & \multicolumn{1}{l}{MSE} & \multicolumn{1}{l}{MAE} & \multicolumn{1}{l}{MSE} & \multicolumn{1}{l}{MAE} & \multicolumn{1}{l}{MSE} & \multicolumn{1}{l}{MAE} & \multicolumn{1}{l}{MSE} & \multicolumn{1}{l}{MAE} & \multicolumn{1}{l}{MSE} & \multicolumn{1}{l}{MAE} & \multicolumn{1}{l}{MSE} & \multicolumn{1}{l}{MAE} & \multicolumn{1}{l}{MSE} & \multicolumn{1}{l}{MAE} \\
        \multirow{5}{*}{\begin{sideways}ETTh1\end{sideways}} & 96 & \textbf{0.368} & \textbf{0.394} & \underline{0.375} & \underline{0.400} & 0.386 & 0.405 & 0.387 & 0.411 & 0.460 & 0.447 & 0.384 & 0.402 & 0.423 & 0.448 & 0.438 & 0.433 & 0.397 & 0.412 \\
         & 192 & \textbf{0.420} & \textbf{0.418} & \underline{0.429} & \underline{0.421} & 0.441 & 0.436 & 0.441 & 0.437 & 0.512 & 0.477 & 0.436 & 0.429 & 0.471 & 0.474 & 0.493 & 0.466 & 0.446 & 0.441 \\
         & 336 & \textbf{0.452} & \textbf{0.433} & \underline{0.484} & \underline{0.458} & 0.487 & \underline{0.458} & 0.507 & 0.467 & 0.546 & 0.496 & 0.638 & 0.469 & 0.570 & 0.546 & 0.547 & 0.495 & 0.489 & 0.467 \\
         & 720 & \textbf{0.449} & \textbf{0.449} & \underline{0.498} & \underline{0.482} & 0.503 & 0.491 & 0.527 & 0.548 & 0.544 & 0.517 & 0.521 & 0.500 & 0.653 & 0.621 & 0.586 & 0.538 & 0.513 & 0.510 \\ \cmidrule{2-20}
         & Avg & \textbf{0.422} & \textbf{0.423} & \underline{0.447} & \underline{0.440} & 0.454 & 0.447 & 0.466 & 0.467 & 0.516 & 0.484 & 0.495 & 0.450 & 0.529 & 0.522 & 0.516 & 0.483 & 0.461 & 0.457 \\ \midrule
        \multirow{5}{*}{\begin{sideways}ETTh2\end{sideways}} & 96 & \textbf{0.282} & \textbf{0.334} & \underline{0.289} & \underline{0.341} & 0.297 & 0.349 & 0.308 & 0.357 & 0.308 & 0.355 & 0.340 & 0.374 & 0.745 & 0.584 & 0.322 & 0.364 & 0.340 & 0.394 \\
         & 192 & \textbf{0.359} & \textbf{0.385} & \underline{0.372} & \underline{0.392} & 0.380 & 0.400 & 0.395 & 0.404 & 0.393 & 0.405 & 0.402 & 0.414 & 0.877 & 0.656 & 0.404 & 0.414 & 0.482 & 0.479 \\
         & 336 & \textbf{0.374} & \textbf{0.404} & \underline{0.386} & \underline{0.414} & 0.428 & 0.432 & 0.428 & 0.434 & 0.427 & 0.436 & 0.452 & 0.452 & 1.043 & 0.731 & 0.435 & 0.445 & 0.591 & 0.541 \\
         & 720 & \textbf{0.405} & \textbf{0.427} & \underline{0.412} & \underline{0.434} & 0.427 & 0.445 & 0.443 & 0.451 & 0.436 & 0.450 & 0.462 & 0.468 & 1.104 & 0.763 & 0.447 & 0.458 & 0.839 & 0.661 \\ \cmidrule{2-20}
         & Avg & \textbf{0.355} & \textbf{0.387} & \underline{0.364} & \underline{0.395} & 0.383 & 0.407 & 0.394 & 0.412 & 0.391 & 0.411 & 0.414 & 0.427 & 0.942 & 0.684 & 0.402 & 0.420 & 0.563 & 0.519 \\ \midrule
        \multirow{5}{*}{\begin{sideways}ETTm1\end{sideways}} & 96 & \textbf{0.314} & \textbf{0.350} & \underline{0.320} & \underline{0.357} & 0.334 & 0.368 & 0.331 & 0.378 & 0.352 & 0.374 & 0.338 & 0.375 & 0.404 & 0.426 & 0.353 & 0.370 & 0.346 & 0.374 \\
         & 192 & \textbf{0.358} & \textbf{0.375} & \underline{0.361} & \underline{0.381} & 0.377 & 0.391 & 0.386 & 0.399 & 0.390 & 0.393 & 0.374 & 0.387 & 0.450 & 0.451 & 0.389 & 0.387 & 0.382 & 0.391 \\
         & 336 & \textbf{0.384} & \textbf{0.395} & \underline{0.390} & \underline{0.404} & 0.426 & 0.420 & 0.426 & 0.421 & 0.421 & 0.414 & 0.410 & 0.411 & 0.532 & 0.515 & 0.421 & 0.408 & 0.415 & 0.415 \\
         & 720 & \textbf{0.448} & \textbf{0.432} & \underline{0.454} & \underline{0.441} & 0.491 & 0.459 & 0.489 & 0.465 & 0.462 & 0.449 & 0.478 & 0.450 & 0.666 & 0.589 & 0.481 & \underline{0.441} & 0.473 & 0.451 \\ \cmidrule{2-20}
         & Avg & \textbf{0.376} & \textbf{0.388} & \underline{0.381} & \underline{0.395} & 0.407 & 0.410 & 0.408 & 0.416 & 0.406 & 0.407 & 0.400 & 0.406 & 0.513 & 0.495 & 0.411 & 0.402 & 0.404 & 0.408 \\ \midrule
        \multirow{5}{*}{\begin{sideways}ETTm2\end{sideways}} & 96 & \textbf{0.171} & \textbf{0.253} & \underline{0.175} & \underline{0.258} & 0.180 & 0.264 & 0.179 & 0.282 & 0.183 & 0.270 & 0.187 & 0.267 & 0.287 & 0.366 & 0.183 & 0.266 & 0.193 & 0.293 \\
         & 192 & \textbf{0.234} & \textbf{0.294} & \underline{0.237} & \underline{0.299} & 0.250 & 0.309 & 0.244 & 0.305 & 0.255 & 0.314 & 0.249 & 0.309 & 0.414 & 0.492 & 0.248 & 0.305 & 0.284 & 0.361 \\
         & 336 & \textbf{0.292} & \textbf{0.333} & \underline{0.298} & \underline{0.340} & 0.311 & 0.348 & 0.320 & 0.357 & 0.309 & 0.347 & 0.321 & 0.351 & 0.597 & 0.542 & 0.309 & 0.343 & 0.382 & 0.429 \\
         & 720 & \textbf{0.387} & \textbf{0.390} & \underline{0.391} & \underline{0.396} & 0.412 & 0.407 & 0.419 & 0.432 & 0.412 & 0.404 & 0.408 & 0.403 & 1.730 & 1.042 & 0.410 & 0.400 & 0.558 & 0.525 \\ \cmidrule{2-20}
         & Avg & \textbf{0.271} & \textbf{0.317} & \underline{0.275} & \underline{0.323} & 0.288 & 0.332 & 0.290 & 0.344 & 0.290 & 0.334 & 0.291 & 0.333 & 0.757 & 0.610 & 0.287 & 0.329 & 0.354 & 0.402 \\ \midrule
        \multirow{5}{*}{\begin{sideways}Weather\end{sideways}} & 96 & \textbf{0.162} & \textbf{0.204} & \underline{0.163} & \underline{0.209} & 0.174 & 0.214 & 0.175 & 0.247 & 0.186 & 0.227 & 0.172 & 0.220 & 0.195 & 0.271 & 0.195 & 0.236 & 0.195 & 0.252 \\
         & 192 & \underline{0.209} & \textbf{0.246} & \textbf{0.208} & \underline{0.250} & 0.221 & 0.254 & 0.224 & 0.294 & 0.234 & 0.265 & 0.219 & 0.261 & \underline{0.209} & 0.277 & 0.239 & 0.271 & 0.237 & 0.295 \\
         & 336 & 0.263 & \textbf{0.287} & \underline{0.251} & \textbf{0.287} & 0.278 & 0.296 & 0.262 & 0.326 & 0.284 & 0.301 & \textbf{0.246} & 0.337 & 0.273 & 0.332 & 0.289 & 0.306 & 0.282 & 0.331 \\
         & 720 & \textbf{0.339} & \textbf{0.338} & \textbf{0.339} & \underline{0.341} & 0.358 & 0.347 & 0.349 & 0.348 & 0.356 & 0.349 & 0.365 & 0.359 & 0.379 & 0.401 & 0.361 & 0.351 & 0.345 & 0.382 \\ \cmidrule{2-20}
         & Avg & \underline{0.243} & \textbf{0.269} & \textbf{0.240} & \underline{0.271} & 0.258 & 0.278 & 0.253 & 0.304 & 0.265 & 0.285 & 0.251 & 0.294 & 0.264 & 0.320 & 0.271 & 0.291 & 0.265 & 0.315 \\ \midrule
        \multirow{5}{*}{\begin{sideways}Electricity\end{sideways}} & 96 & \underline{0.150} & \underline{0.241} & 0.153 & 0.247 & \textbf{0.148} & \textbf{0.240} & 0.190 & 0.299 & 0.190 & 0.296 & 0.168 & 0.272 & 0.219 & 0.314 & 0.198 & 0.274 & 0.210 & 0.302 \\
         & 192 & \textbf{0.162} & \textbf{0.252} & 0.166 & 0.256 & \textbf{0.162} & \underline{0.253} & 0.216 & 0.323 & 0.199 & 0.304 & 0.184 & 0.322 & 0.231 & 0.322 & 0.198 & 0.278 & 0.210 & 0.305 \\
         & 336 & \underline{0.179} & \underline{0.270} & 0.185 & 0.277 & \textbf{0.178} & \textbf{0.269} & 0.226 & 0.334 & 0.217 & 0.319 & 0.198 & 0.300 & 0.246 & 0.337 & 0.217 & 0.300 & 0.223 & 0.319 \\
         & 720 & \textbf{0.217} & \textbf{0.304} & 0.225 & \underline{0.310} & 0.225 & 0.317 & 0.250 & 0.353 & 0.258 & 0.352 & 0.220 & 0.320 & 0.280 & 0.363 & 0.278 & 0.356 & 0.258 & 0.350 \\ \cmidrule{2-20}
         & Avg & \textbf{0.177} & \textbf{0.267} & 0.182 & 0.272 & \underline{0.178} & \underline{0.270} & 0.220 & 0.327 & 0.216 & 0.318 & 0.193 & 0.304 & 0.244 & 0.334 & 0.223 & 0.302 & 0.225 & 0.319 \\ \midrule
        \multirow{5}{*}{\begin{sideways}Traffic\end{sideways}} & 96 & 0.465 & 0.286 & \underline{0.462} & \underline{0.285} & \textbf{0.395} & \textbf{0.268} & 0.499 & 0.344 & 0.526 & 0.347 & 0.593 & 0.321 & 0.644 & 0.429 & 0.647 & 0.384 & 0.650 & 0.396 \\
         & 192 & 0.475 & \underline{0.290} & \underline{0.473} & 0.296 & \textbf{0.417} & \textbf{0.276} & 0.540 & 0.370 & 0.522 & 0.332 & 0.617 & 0.336 & 0.665 & 0.431 & 0.600 & 0.361 & 0.598 & 0.370 \\
         & 336 & \underline{0.489} & \underline{0.296} & 0.498 & \underline{0.296} & \textbf{0.433} & \textbf{0.283} & 0.557 & 0.378 & 0.517 & 0.334 & 0.629 & 0.336 & 0.674 & 0.420 & 0.610 & 0.367 & 0.605 & 0.373 \\
         & 720 & 0.527 & 0.318 & \underline{0.506} & \underline{0.313} & \textbf{0.467} & \textbf{0.302} & 0.586 & 0.397 & 0.552 & 0.352 & 0.640 & 0.350 & 0.683 & 0.424 & 0.691 & 0.425 & 0.645 & 0.394 \\ \cmidrule{2-20}
         & Avg & 0.489 & \underline{0.297} & \underline{0.484} & \underline{0.297} & \textbf{0.428} & \textbf{0.282} & 0.546 & 0.372 & 0.529 & 0.341 & 0.620 & 0.336 & 0.667 & 0.426 & 0.637 & 0.384 & 0.625 & 0.383 \\ \midrule
        \multicolumn{2}{r}{1st Cnt:} & \multicolumn{1}{c}{25} & \multicolumn{1}{c}{28} & 3 & 1 & 8 & 7 & 0 & 0 & 0 & 0 & 1 & 0 & 0 & 0 & 0 & 0 & 0 & 0 \\ \bottomrule
    \end{tabularx}
    \caption{Multivariate long-term time series forecasting results under the unified setting of TimeMixer, including similar look-back window, batch size, and epochs. We optimize other hyperparameters including learning rate, $\psi$, $m$, $t_f$, $d_f$, dropouts, Patch, Stride, and $d$. The results of the models marked with $(*)$ are taken from the corresponding original papers. Other models' results are taken from TimeMixer \cite{c:timemixer}.}
    \label{tab:unified_result}
\end{table*}

\subsection{Univariate Long-term forecasting result}
In addition to the superior performance of our model in multivariate long-term time series forecasting, it also achieves excellent results in the univariate long-term forecasting task. The detailed results on the ETT datasets are presented in Table \ref{tab:univariate_result}. Across all prediction lengths, WPMixer outperforms PatchTST, which is the best-performing method in the univariate forecasting tasks, and other existing methods.

\begin{table*}[!htb]
    \centering
    \fontsize{9}{6}\selectfont
    \begin{tabularx}{\textwidth}{cccccccccccccccccc}
        \toprule
        \multicolumn{1}{l}{} & \multicolumn{1}{l}{} & \multicolumn{2}{c}{WPMixer} & \multicolumn{2}{c}{PatchTST/64} & \multicolumn{2}{c}{PatchTST/42} & \multicolumn{2}{c}{Dlinear} & \multicolumn{2}{c}{FEDformer} & \multicolumn{2}{c}{Autoformer} & \multicolumn{2}{c}{Informer} & \multicolumn{2}{c}{LogTrans} \\ \midrule
        Data & Pred. len & MSE & MAE & MSE & MAE & MSE & MAE & MSE & MAE & MSE & MAE & MSE & MAE & MSE & MAE & MSE & MAE \\ \midrule
        \multirow{5}{*}{ETTh1} & 96 & \textbf{0.055} & 0.181 & 0.059 & 0.189 & \textbf{0.055} & \textbf{0.179} & 0.056 & \underline{0.180} & 0.079 & 0.215 & 0.071 & 0.206 & 0.193 & 0.377 & 0.283 & 0.468 \\
         & 192 & \textbf{0.065} & \textbf{0.200} & 0.074 & 0.215 & \underline{0.071} & 0.205 & \underline{0.071} & \underline{0.204} & 0.104 & 0.245 & 0.114 & 0.262 & 0.217 & 0.395 & 0.234 & 0.409 \\
         & 336 & \textbf{0.071} & \textbf{0.212} & \underline{0.076} & \underline{0.220} & 0.081 & 0.225 & 0.098 & 0.244 & 0.119 & 0.270 & 0.107 & 0.258 & 0.202 & 0.381 & 0.386 & 0.546 \\
         & 720 & \textbf{0.083} & \textbf{0.229} & \underline{0.087} & 0.236 & \underline{0.087} & \underline{0.232} & 0.189 & 0.359 & 0.142 & 0.299 & 0.126 & 0.283 & 0.183 & 0.355 & 0.475 & 0.629 \\ \cmidrule{2-18}
         & Avg. & \textbf{0.068} & \textbf{0.205} & \underline{0.074} & 0.215 & \underline{0.074} & \underline{0.210} & 0.104 & 0.247 & 0.111 & 0.257 & 0.105 & 0.252 & 0.199 & 0.377 & 0.345 & 0.513 \\  \midrule
        \multirow{5}{*}{ETTh2} & 96 & \textbf{0.122} & \underline{0.278} & 0.131 & 0.284 & 0.129 & 0.282 & 0.131 & 0.279 & \underline{0.128} & \textbf{0.271} & 0.153 & 0.306 & 0.213 & 0.373 & 0.217 & 0.379 \\
         & 192 & \textbf{0.163} & \textbf{0.324} & 0.171 & 0.329 & \underline{0.168} & \underline{0.328} & 0.176 & 0.329 & 0.185 & 0.330 & 0.204 & 0.351 & 0.227 & 0.387 & 0.281 & 0.429 \\
         & 336 & \textbf{0.159} & \textbf{0.324} & \underline{0.171} & \underline{0.336} & 0.185 & 0.351 & 0.209 & 0.367 & 0.231 & 0.378 & 0.246 & 0.389 & 0.242 & 0.401 & 0.293 & 0.437 \\
         & 720 & \textbf{0.197} & \textbf{0.355} & \underline{0.223} & \underline{0.380} & 0.224 & 0.383 & 0.276 & 0.426 & 0.278 & 0.420 & 0.268 & 0.409 & 0.291 & 0.439 & 0.218 & 0.387 \\ \cmidrule{2-18}
         & Avg. & \textbf{0.160} & \textbf{0.320} & \underline{0.174} & \underline{0.332} & 0.177 & 0.336 & 0.198 & 0.350 & 0.206 & 0.350 & 0.218 & 0.364 & 0.243 & 0.400 & 0.252 & 0.408 \\  \midrule
        \multirow{5}{*}{ETTm1} & 96 & \textbf{0.026} & \textbf{0.121} & \textbf{0.026} & 0.123 & \textbf{0.026} & \textbf{0.121} & 0.028 & 0.123 & 0.033 & 0.140 & 0.056 & 0.183 & 0.109 & 0.277 & 0.049 & 0.171 \\
         & 192 & \textbf{0.038} & \textbf{0.149} & 0.040 & 0.151 & \underline{0.039} & \underline{0.150} & 0.045 & 0.156 & 0.058 & 0.186 & 0.081 & 0.216 & 0.151 & 0.310 & 0.157 & 0.317 \\
         & 336 & \textbf{0.051} & \textbf{0.172} & \underline{0.053} & 0.174 & \underline{0.053} & \underline{0.173} & 0.061 & 0.182 & 0.084 & 0.231 & 0.076 & 0.218 & 0.427 & 0.591 & 0.289 & 0.459 \\
         & 720 & \textbf{0.068} & \textbf{0.197} & \underline{0.073} & \underline{0.206} & 0.074 & 0.207 & 0.080 & 0.210 & 0.102 & 0.250 & 0.110 & 0.267 & 0.438 & 0.586 & 0.43 & 0.579 \\  \cmidrule{2-18}
         & Avg. & \textbf{0.046} & \textbf{0.160} & \underline{0.048} & 0.164 & \underline{0.048} & \underline{0.163} & 0.054 & 0.168 & 0.069 & 0.202 & 0.081 & 0.221 & 0.281 & 0.441 & 0.231 & 0.382 \\  \midrule
        \multirow{5}{*}{ETTm2} & 96 & \textbf{0.063} & \underline{0.184} & 0.065 & 0.187 & 0.065 & 0.186 & \textbf{0.063} & \textbf{0.183} & 0.067 & 0.198 & 0.065 & 0.189 & 0.088 & 0.225 & 0.075 & 0.208 \\
         & 192 & \underline{0.093} & \underline{0.229} & \underline{0.093} & 0.231 & 0.094 & 0.231 & \textbf{0.092} & \textbf{0.227} & 0.102 & 0.245 & 0.118 & 0.256 & 0.132 & 0.283 & 0.129 & 0.275 \\
         & 336 & \textbf{0.118} & \underline{0.263} & 0.121 & 0.266 & 0.120 & 0.265 & \underline{0.119} & \textbf{0.261} & 0.130 & 0.279 & 0.154 & 0.305 & 0.180 & 0.336 & 0.154 & 0.302 \\
         & 720 & \textbf{0.166} & \textbf{0.318} & 0.172 & 0.322 & \underline{0.171} & 0.322 & 0.175 & \underline{0.320} & 0.178 & 0.325 & 0.182 & 0.335 & 0.300 & 0.435 & 0.16 & 0.321 \\  \cmidrule{2-18}
         & Avg. & \textbf{0.110} & \underline{0.249} & 0.113 & 0.252 & 0.113 & 0.251 & \underline{0.112} & \textbf{0.248} & 0.119 & 0.262 & 0.130 & 0.271 & 0.175 & 0.320 & 0.130 & 0.277 \\ \bottomrule
    \end{tabularx}
    \caption{Univariate long-term time series prediction results. The results of the other models are taken from PatchTST \cite{c:patchtst}.}
    \label{tab:univariate_result}
\end{table*}

\end{document}